\documentclass[english,11pt]{article}

\usepackage{times}
\usepackage{babel}
\usepackage{latexsym}
\usepackage{float}
\usepackage{amssymb}
\usepackage{amsmath}

\newcommand{\lar}{\leftarrow}

\newcommand{\GL}{\mathit{GL}}

\newcommand{\at}{\mathit{At}}

\newcommand{\CCGL}{\mathit{CCGL}}
\newcommand{\cc}{\mathit{CC}}
\newcommand{\lfp}{\mathit{lfp}}
\newcommand{\rar}{\rightarrow}

\newcommand{\la}{\langle}
\newcommand{\ra}{\rangle}

\newcommand{\posBody}{\mathit{posBody}}
\newcommand{\negBody}{\mathit{negBody}}

\newtheorem{proposition}{Proposition}[section]

\newtheorem{example}{Example}[section]


\begin{document}

\title{An Application of Proof-Theory in Answer Set Programming}

\author{V. W. Marek
Department of Computer Science\\ 
University of Kentucky\\ 
Lexington, KY 40506
\and
J.B. Remmel\\
Departments of Computer Science and Mathematics\\
University of California\\
La Jolla, CA 92093}

\maketitle

\vspace*{.5in}
\begin{abstract}
Using a characterization of stable models of logic programs $P$ as
satisfying valuations of a suitably chosen propositional theory, 
called the set of {\em reduced defining equations} ${r\Phi}_P$, we 
show that the finitary character of that theory ${r\Phi}_P$ is
equivalent to a certain continuity property of the Gelfond-Lifschitz
operator  $\GL_P$ associated with the program $P$. \\
We discuss
possible extensions of techniques proposed in this paper to the context of
cardinality constraints.

\end{abstract}

\section{Introduction}\label{intro}

The use of proof theory in logic based formalisms for constraint solving 
is pervasive. For example, in Satisfiability (SAT), proof theoretic methods 
are used to find lower bounds on complexity of various SAT algorithms. 
However, proof-theoretic 
methods have not played as prominent role in Answer Set Programming (ASP) 
formalisms. This is not to say that there were no attempts
to apply proof-theoretic methods in ASP. To give a few examples, Marek
and Truszczynski in \cite{mt93} used the proof-theoretic methods to
characterize Reiter's extensions in Default Logic (and thus stable
semantics of logic programs). Bonatti \cite{bo04} and separately
Milnikel \cite{mi05} devised non-monotonic proof systems to study
skeptical consequences of programs and default theories. Lifschitz
\cite{li96} used proof-theoretic methods to approximate
well-founded semantics of logic programs. Bondarenko et.al. \cite{btk93}
studied an approach to stable semantics  using methods with a clear 
proof-theoretic flavor. Marek, Nerode, and Remmel in a series of papers, 
\cite{MNR1,MNR2,MNR3,MNR4,MNR5,MNR6}, developed proof theoretic 
methods to study what they termed {\it non-monotonic rule systems} 
which have as special cases almost all ASP formalisms that have 
been seriously studied in the literature.  
Recently the area of proof systems
for ASP (and more generally, nonmonotonic logics) received a lot of attention
\cite{gs07,jo07}. It is clear that the community feels that an additional
research of this area is necessary.
Nevertheless, there is no clear classification of proof systems
for nonmonotonic reasoning analogous to that present in classical logic,
and SAT in particular.

In this paper, we define a notion of $P$-proof schemes, 
which is a kind of a proof system that was previously used by Marek, Nerode, 
and Remmel to study complexity issues for stable semantics of logic 
programs \cite{MNR5}. 
This proof system abstracts of
$M$-proofs of \cite{mt93} and produces  Hilbert-style proofs. The
nonmonotonic character of our $P$-proofs is provided by the presence of
guards, called the {\em support} of the proof scheme,  
to insure context-dependence. A different but equivalent, presentation 
of proof schemes, using a guarded resolution is also possible \cite{mr09}.

We shall show that we can use $P$-proof schemes to find a 
characterization of stable models via
{\em reduced defining equations}. While in general these defining
equations may be infinite, we study  the case of programs for which
all these equations are finite. This resulting class of programs, 
called FSP-programs, turn out to be characterized by 
a form of continuity of the Gelfond-Lifschitz operator. 


\subsection{Contributions of the paper}

The contributions of this paper consist, primarily of investigations that
elucidate the proof-theoretical character of the stable semantics for logic
programs, an area with 20 years history \cite{gl88}. The principal results of
this paper are:
\begin{enumerate}
\item We show that the Gelfond-Lifschitz operator $\mathit{GL}_P$ is, in fact a
proof-theoretical construct (Proposition \ref{p.gl})
\item As a result of the analysis of the Gelfond-Lifschitz operator we are able
to show that the upper-half continuity of that operator is equivalent to
finiteness of (propositional) formulas in a certain class  associated with the
program $P$ (Proposition \ref{p.fsp})
\end{enumerate}
We also discuss possible extension of these results to the case of programs
with cardinality constraints.

\section{Preliminaries}\label{prelim}

Let $\at$ be a countably infinite set of atoms.
We will study programs consisting of clauses built of the
atoms from $\at$. A {\em program clause} $C$ is a string of
the form
\begin{equation}\label{clause}
p \lar q_1,\ldots, q_m, \neg r_1,\ldots, \neg r_n
\end{equation}
The integers $m$ or $n$ or both can be $0$. The
atom $p$ will be called the head of $C$ and denoted $\mathit{head}
(C)$. We let $\posBody (C)$ denote the set $\{q_1,\ldots, q_m\}$ and
$\negBody (C)$ denote the set $\{r_1,\ldots, r_n\}$.  For any set of
atoms $X$, we let $\neg X$
denote the conjunction of negations of atoms from $X$. Thus,  we can write
clause (\ref{clause}) as 
\[
\mathit{head}(C)  \lar \posBody(C), \neg \negBody(C).
\]
Let us stress that the set $\negBody(C)$ is a set of
atoms, not a set of negated atoms as is sometimes used in the
literature.  A normal propositional program is a set $P$ of such
clauses. For any $M \subseteq \at$, we say that $M$ is model of $C$ if 
whenever $q_1, \ldots, q_m \in M$ and $\{r_1, \ldots, r_n\} 
\cap M = \emptyset$, then $p \in M$. We say that $M$ is a model of 
a program $P$ if $M$ is a model of each clause $C \in P$.  Horn clauses 
are clauses with no negated literals, i.e. clauses of the 
form (\ref{clause}) where $n=0$. We will denote by 
$\mathit{Horn}(P)$ the part of the program
$P$ consisting of its Horn clauses.
Horn programs are logic programs $P$ consisting entirely of Horn 
clauses. Thus for a Horn program $P$, $P = \mathit{Horn}(P)$.

Each Horn program $P$ has a least model over the Herbrand base and 
the least model of $P$ is  
the least fixed point of a continuous operator $T_P$ representing one-step
Horn clause logic deduction (\cite{ll89}). That is, for any set 
$I \subseteq \at$, we let 
$T_{P} (I)$  equal the set of all $p \in \at$ such that 
there is a clause $C = p\leftarrow q_1,\ldots, q_m$ in $P$ and $q_1, \ldots,
q_m \in I$. Then $T_P$ has a least fixed point $F_P$ which is obtained by
iterating $T_P$ starting at the empty set for $\omega$ steps, 
i.e., $F_P = \bigcup_{ n \in \omega} T^n_P(\emptyset)$
where for any $I \subseteq \at$, $T^0_P(I) =I$ and 
$T^{n+1}_P(I) = T_P(T^n_P(I))$. Then $F_P$ is the least model of $P$.

The semantics of interest for us is the {\em stable semantics} of normal
programs, although we will discuss some extensions in Section \ref{ext}.
The stable models of a program $P$ are defined as fixed points of the
operator $T_{P,M}$. This operator is defined on the set of all subsets
of $\at$, ${\cal P}(\at )$. If $P$ is a program and 
$M\subseteq \at$ is a subset of the
Herbrand base, define operator $T_{P,M} \colon
{\cal P}(\at)\rightarrow {\cal P}(\at)$ as follows:
\begin{multline*}
T_{P,M} (I) = \{ p\colon {\rm there\ exist\ a\ clause\ }
C = p\leftarrow q_1 ,\ldots ,q_m,\neg r_1 ,\ldots , \neg r_n \\
{\rm in\ } P \ \ {\rm such \ that\ }
q_1\in I,\ldots , q_m\in I, r_1\notin M, \ldots, r_n \notin M\}
\end{multline*}
The following is immediate, see \cite{ap90} for unexplained notions.
\begin{proposition}
For every program $P$ and every set $M$ of atoms
the operator $T_{P,M}$ is monotone and continuous.
\end{proposition}
Thus the operator $T_{P,M}$ 
like all monotonic continuous operators,
possesses a least fixed point $F_{P,M}$. 

Given program $P$  and
$M\subseteq \at$, we define the {\em Gelfond-Lifschitz reduct} of $P$, $P_M$,
as follows. For every clause $C = p\leftarrow q_1,\ldots, q_m, 
\neg r_1, \ldots, \neg r_n$ of $P$, execute 
the following operations.\\
(1) If some atom $r_i$, $1\le i\le n$, belongs to $M$, then eliminate $C$
altogether. \\ 
(2) In the remaining clauses that have not been eliminated
by operation (1), eliminate all the negated atoms.\\
The resulting program $P_{M}$ is a Horn propositional
program. The program  $P_{M}$ possesses a least
Herbrand model. If that least model 
of $P_M$ coincides with $M$, then $M$ is
called a {\em stable model} for $P$. This gives rise to an 
operator $GL_P$ which associates to each $M \subseteq \at$, 
the least fixed point of $T_{P,M}$.  We will discuss the operator 
$GL_P$ and its proof-theoretic connections in section \ref{gl}.

\section{Proof schemes and reduced defining equations}\label{psr}

In this section we recall the notion of a {\em proof scheme} as 
defined in \cite{MNR1,mt93} 
and introduce a related notion of {\em defining equations}.

Given a propositional logic program $P$, a proof scheme is defined
by induction on its length. Specifically, a  proof scheme w.r.t. $P$ (in
short $P$-proof scheme) is a sequence
$S = \la \la C_1,p_1\ra,\ldots, \la C_n , p_n \ra, U \ra$
subject to the following conditions:\\
(I) when $n = 1$, $\la \la C_1, p_1 \ra ,U\ra$ is a $P$-proof scheme
if $C_1 \in P$,  $p_1 = \mathit{head}(C_1)$, $\posBody(C_1) = 
\emptyset$, and $U = \negBody(C_1)$ and \\
(II) when
$\la \la C_1,p_1\ra,\ldots, \la C_n , p_n \ra, U \ra$ is a $P$-proof
scheme, \\
$C = p \leftarrow \posBody(C), \neg \negBody(C)$
is a clause in the program  $P$, and $\posBody(C) \subseteq
\{p_1,\ldots,p_n\}$, then
\[
\la \la C_1,p_1\ra,\ldots, \la C_n , p_n \ra,  \la C, p\ra,  U \cup
\negBody(C) \ra
\]
is a $P$-proof scheme.\\
When  $ S = \la \la C_1,p_1\ra,\ldots, \la C_n , p_n \ra, U \ra$ is a $P$-proof
scheme,  then we call (i) the integer $n$ -- the {\em length} of  $S$, 
(ii) the set $U$ -- the {\em  support} of $S$, and (iii) the atom 
$p_n$ -- the {\em conclusion} of $S$. 
We denote $U$ by $\mathit{supp}(S)$.

\begin{example}\label{ex1}
{\rm 
Let $P$ be a program consisting of four clauses:
$ C_1 = p \leftarrow$,
$ C_2 = q \leftarrow p,\neg r$,
$ C_3 = r \leftarrow \neg q$, and
$ C_4 = s \leftarrow \neg t$.
Then we have the following examples of $P$-proof schemes:
\begin{enumerate}
\item[(a)] $\la\la C_1,p\ra,\emptyset\ra$ is a $P$-proof scheme of length
$1$ with conclusion $p$ and empty support.
\item[(b)] $\la\la C_1,p\ra,\la C_2, q\ra ,\{ r\}\ra$ is a $P$-proof scheme of 
length $2$ with conclusion $q$ and support $\{r\}$.
\item[(c)] $\la\la C_1,p\ra,\la C_3, r\ra ,\{ q\}\ra$ is a $P$-proof scheme of 
length $2$ with conclusion $r$  and support $\{q\}$.
\item[(d)] $\la\la C_1,p\ra, \la C_2, q \ra , \la C_3, r\ra ,\{ q,r\}\ra$ 
is a $P$-proof scheme of 
length $3$ with conclusion $r$ and support $\{q,r\}$.
\end{enumerate}
Proof scheme in (c) is an example of a proof scheme with unnecessary
items (the first term). Proof scheme (d) is an example 
of a proof scheme which is not internally consistent in 
that $r$ is in the support of its proof scheme and is also its 
conclusion. $\hfill\Box$
}
\end{example}

A $P$-proof scheme carries within itself its own applicability condition. In
effect, a $P$-proof scheme is a {\em conditional} proof of its
conclusion. It becomes applicable when all the constraints collected in
the support are satisfied. Formally, for any set of atoms $M$, we say 
that a $P$-proof scheme $S$ is $M$-{\em applicable} if $M \cap
\mathit{supp}(S) = \emptyset$. We also say that $M$ {\em admits} $S$ if
$S$ is $M$-applicable.

The fundamental connection between proof schemes and stable models 
\cite{MNR1,mt93} is given by the following proposition.
\begin{proposition}\label{p.char}
For every normal propositional program $P$ and every set $M$ of
atoms, $M$ is a stable model of $P$ if and only if the following 
conditions hold.
\begin{enumerate}
\item[(i)] For every $p \in M$,  there is a $P$-proof scheme $S$ with
conclusion $p$ such
that $M$ admits $S$.
\item[(ii)] For every $p \notin M$,  there is no $P$-proof scheme $S$ 
with conclusion $p$ such that $M$ admits $S$.
\end{enumerate}
\end{proposition}

Proposition \ref{p.char} says that the presence and absence of the atom 
$p$ in a stable
model depends {\em only} on the supports of proof schemes. 
This fact naturally leads to a characterization of stable models 
in terms of propositional satisfiability. Given $p \in \at$, 
the {\em defining equation} for $p$ w.r.t. $P$ is the
following propositional formula:
\begin{equation}\label{defeq}
p \Leftrightarrow (\neg U_1 \lor \neg U_2 \lor \ldots )
\end{equation}
where $\la U_1,U_2,\ldots \ra$ is the list of all supports of $P$-proof
schemes. Here for any finite set $S = \{s_1, \ldots, s_n\}$ of atoms, 
$\neg S = \neg s_1 \wedge \cdots \wedge \neg s_n.$
If $p$ is not the conclusion of any proof scheme, then 
we set the defining equation of $p$ to be $p \Leftrightarrow \bot$. 
In the case, where all the supports of proof schemes of $p$ are 
empty, we set the defining equation of $p$ to be $p \Leftrightarrow \top$. 
Up to a total ordering of the finite sets of atoms such a formula is
unique. For example, suppose we fix a total order on $\at$, 
$p_1 < p_2 < \cdots $. Then given two sets of atoms, 
$U = \{u_1 < \cdots < u_m\}$ and $V = \{v_1 < \cdots < 
v_n\}$, we say that $U \prec V$, if either 
(i) $u_m < v_n$, (ii) $u_m = v_n$ and $m < n$, or (iii) $u_m = v_n$, 
$n= m$, and $(u_1, \ldots, u_n)$ is lexicographically 
less than $(v_1, \ldots, v_n)$.  We say that (\ref{defeq}) 
is the {\it defining equation}
for $p$ relative to $P$ if $U_1 \prec U_2 \prec \cdots $. We will 
denote  the defining equation for
$p$ with respect to $P$ by $\mathit{Eq}_p^P$. 

For example, if $P$ is a Horn program, then for every atom $p$, either the 
support of all its proof schemes are empty or  $p$ is not the conclusion of any
proof scheme. The first of these alternatives occurs when $p$ belongs to the
least model of $P$, $\mathit{lm}(P)$. The second alternative occurs when $p
\notin \mathit{lm}(P)$. The defining equations are $p \Leftrightarrow \top$
(that is $p$) when $p \in \mathit{lm}(P)$ and 
$p \Leftrightarrow \bot$ (that is $\neg p$) when $p \notin \mathit{lm}(P)$. 
When $P$ is a stratified program the defining equations are more complex, but
the resulting theory is logically equivalent to
\[
\{ p : p \in \mathit{Perf}_P \} \cup
\{ \neg p : p \notin \mathit{Perf}_P \}
\]
where $\mathit{Perf}_P$ is the unique stable model of $P$.

Let $\Phi_P$ be the set $\{\mathit{Eq}_p^P : p
\in \at\}$. We then have the following consequence of Proposition
\ref{p.char}.
\begin{proposition}\label{p.eq}
Let $P$ be a normal propositional program.
Then stable models of $P$ are precisely the propositional models of the
theory $\Phi_P$.
\end{proposition}
When $P$ is {\em purely negative}, i.e. all clauses $C$  of $P$ have 
$\mathit{PosBody}(C) = \emptyset$, the stable and supported
models of $P$ coincide  \cite{dk89} and the defining equations reduce to
Clark's completion \cite{cl78} of $P$.

Let us observe that in general the propositional formulas on the
right-hand-side of the defining equations may be infinite.

\begin{example}\label{e.2}
{\rm 
Let $P$ be an infinite program consisting of clauses $p \lar \neg p_i$,
for all $i \in n$. In this case, the defining equation for $p$ in $P$ 
is infinite. That is, 
it is
\[
p \Leftrightarrow (\neg p_1 \lor \neg p_2 \lor \neg p_3 \lor \ldots )
\]
$\mbox{ }\hfill\Box$
}
\end{example}
The following observation is quite useful. If $U_1, U_2$ are two finite
sets of propositional atoms then
\[
U_1 \subseteq U_2 \ \mbox{if and only if\ }
\ \neg U_2 \models \neg U_1
\]
Here $\models$ is the propositional consequence relation. The effect
of this observation is that not all the supports of proof schemes are
important, only the inclusion-minimal ones.

\begin{example}\label{ex.3}
{\rm 
Let $P$ be an infinite program consisting of clauses $p \lar 
\neg p_1, \ldots ,\neg p_i$,
for all $i \in N$. The defining equation for $p$ in $P$ is 
\[
p \Leftrightarrow [\neg p_1 \lor (\neg p_1 \land \neg p_2)
 \lor (\neg p_1 \land \neg p_2 \land \neg p_3) \lor \ldots\ ]
\]
which is infinite.  But our observation above implies that this formula is {\em
equivalent} to the formula
\[
p \Leftrightarrow \neg p_1 
\]
$\mbox{ }\hfill\Box$
}
\end{example}

Motivated by the Example \ref{ex.3}, we define the {\em reduced
defining equation} for $p$ relative to $P$ to be the formula
\begin{equation}\label{reddefeq}
p \Leftrightarrow (\neg U_1 \lor \neg U_2 \lor \ldots )
\end{equation}
where $U_i$ range over {\em inclusion-minimal} supports of $P$-proof
schemes for the atom $p$ and $U_1 \prec U_2 \prec \cdots$. 
Again, if $p$ is not the conclusion of any proof scheme, then 
we set the defining equation of $p$ to be $p \Leftrightarrow \bot$. 
In the case, where there is a proof scheme of $p$ with empty support, 
then we set the defining equation of $p$ to be $p \Leftrightarrow \top$. 
We denote this formula as 
$\mathit{rEq}_p^P$, and define $r\Phi_P$ to be the theory consisting of
$\mathit{rEq}_p^P$ for all $p \in \at$. We then have the following
strengthening of Proposition \ref{p.eq}.
\begin{proposition}\label{p.req}
Let $P$ be a normal propositional program.
Then stable models of $P$ are precisely the propositional models of the
theory $r\Phi_P$.
\end{proposition}
In our example \ref{ex.3}, the theory $\Phi_P$ involved formulas with 
infinite disjunctions, but the
theory $r\Phi_P$ contains only normal finite propositions.

Given a normal propositional program $P$, we say that $P$ is a 
{\it finite support program} (FSP-program) 
if all the reduced defining equations for atoms with
respect to $P$ are finite propositional formulas. Equivalently, a
program $P$ is an {\em FSP}-program if for every atom $p$ there is
only finitely many inclusion-minimal supports of $P$-proof schemes for
$p$.

\section{Continuity properties of operators and proof schemes}\label{cont}

In this section we investigate continuity properties of operators and we
will see that one of those properties characterizes the class of FSP programs.

\subsection{Continuity properties of monotone and antimonotone
operators}

Let us recall that ${\cal P}(\at )$ denotes the set of all subsets of $\at$.  
We say that any function $O: {\cal P}(\at ) \rar {\cal P}(\at )$ is 
an operator on the set $\at$ of propositional atoms. 
An operator $O$ is
{\em monotone} if for all sets $X, Y \subseteq \at$, $X
\subseteq Y$ implies $O(X) \subseteq O(Y)$. Likewise 
an operator $O$ is
{\em antimonotone} if for all sets $X, Y \subseteq \at$, $X
\subseteq Y$ implies $O(Y) \subseteq O(X)$. 
For a sequence $\la X_n\ra_{n \in N}$ of sets of atoms, we say that
$\la X_n\ra_{n \in N}$ is {\em monotonically increasing} if for all $i,j
\in  N$, $i \le j$ implies $X_i \subseteq X_j$ and we say that
$\la X_n\ra_{n \in N}$ is {\em monotonically decreasing} if for all $i,j
\in  N$, $i \le j$ implies $X_j \subseteq X_i$. 

There are four distinct classes of operators 
that we shall consider in this paper.  
First, we shall consider two types of monotone operators, upper-half 
continuous monotone operators and lower-half  continuous monotone operators.
That is, we say that a monotone operator $O$ is {\em upper-half continuous} if
for every monotonically increasing sequence $\la X_n\ra_{n \in N}$,
$O(\bigcup_{n \in N} X_n) = \bigcup_{n\in N}O(X_n).$ We say 
that a monotone operator $O$ is {\em lower-half continuous} if for every
monotonically decreasing sequence $\la X_n\ra_{n \in N}$,
$O(\bigcap_{n \in N} X_n) = \bigcap_{n\in N}O(X_n).$
In the Logic Programming literature the first of these properties is
called {\em continuity}. The classic result due to van Emden and Kowalski
is the following.
\begin{proposition}\label{p.vek}
For every Horn program $P$, the operator $T_P$ is upper-half continuous.
\end{proposition}

In general, the operator $T_P$ for Horn programs is {\em not} lower-half
continuous. For example, let $P$ be the program 
consisting of the clauses $p \lar p_i$ for $i \in N$. Then 
the operator $T_P$ is not lower-half continuous. That is, 
if $X_i =\{p_i, p_{i+1}, \ldots \}$, then  clearly 
$p \in T_P(X_i)$ for all $i$. However, $\bigcap_{i} X_i = \emptyset$ and 
$p \not \in T_P(\emptyset)$. 

Lower-half continuous monotone operators have appeared in the Logic Programming
literature \cite{doe94}. Even more generally, for a monotone operator
$O$, let us define its {\em dual} operator $O^d$ as follows:
\[
O^d (X) = \at \setminus O(\at \setminus X).
\]
Then an operator $O$ is upper-half continuous if and only if $O^d$ is
lower-half continuous \cite{jt51}. Therefore, for any Horn program $P$,
the operator $T_P^d$ is lower-half continuous.

In case of antimonotone operators, we have two additional notions of
continuity. We say an antimonotone operator $O$ is {\em upper-half} continuous if for every
monotonically increasing sequence $\la X_n\ra_{n \in N}$,
$O(\bigcup_{n \in N} X_n) = \bigcap_{n\in N}O(X_n).$ Similarly, we say 
an antimonotone operator $O$ is {\em lower-half} continuous if for every
monotonically decreasing sequence $\la X_n\ra_{n \in N}$,
$O(\bigcap_{n \in N} X_n) = \bigcup_{n\in N}O(X_n)$. 

\subsection{Gelfond-Lifschitz operator $\GL_P$ and proof-schemes}\label{gl}
For the completeness sake, let us recall that the Gelfond-Lifschitz operator
for a program $P$ which we denote $GL_P$, assigns to a set of atoms $M$ 
the least fixpoint of of the operator $T_{P,M}$ or, equivalently, the least
model $N_M$ of the program $P_M$ which is the Gelfond-Lifschitz reduct of
$P$ via $M$ \cite{gl88}. The following fact is crucial.

\begin{proposition}[\cite{gl88}] \label{p.anti}
The operator $\GL$ is antimonotone.
\end{proposition}

Here is a useful proof-theoretic characterization of the operator $\GL_P$.
\begin{proposition}\label{p.gl}
Let $P$ be a normal propositional program and $M$ be a set of
atoms. Then 
\begin{multline*}
\GL_P(M) = \{p : \ \mbox{there exists a $P$-proof scheme $S$}
\ \mbox{ such that $M$ admits $S$},  \\
\mbox{and $p$ is the conclusion of $S$}\}
\end{multline*}
\end{proposition}
Proof: Let us assume that  $p \in \GL_P(M)$ that is $p \in N_M$. As $N_M$ 
is the least model of the Horn program $P_M$,  $N_M = \bigcup_{n\in
N}T_{P_M}^n (\emptyset )$. Then it is easy to prove by induction on 
$n$, that if  $p \in T_{P_M}^n (\emptyset )$, then there is a $P$-proof
scheme $S_p$ such that  $p$ is the conclusion of $S_p$ and $S_p$ is
admitted by $M$. Conversely, we can show,  
by induction on the length of the $P$-proof schemes, 
that whenever such $P$-proof scheme $S$ is admitted by $M$, then $p$
belongs to $\GL_P(M)$. $\hfill\Box$

\subsection{Continuity properties of the operator $\GL_P$}

This section will be devoted to proving results on the continuity 
properties of the operator $\GL_P$. First, we prove that for
every program $P$, the operator $\GL_P$ is lower-half continuous.
We then show that if $f$ is a lower-half continuous antimonotone operator, 
then  $f = \GL_P$ for a suitably chosen program
$P$. Finally, we show that the operator $\GL_P$ is upper-half continuous if
and only if $P$ is an {\em FSP}-program. That is, $\GL_P$ is upper-half
continuous if for all atoms $p$ the reduced defining equation 
for any $p$ (w.r.t. $P$) is finite.

\begin{proposition}\label{p.lhcont}
For every normal program $P$, the operator $\GL_P$ is lower-half
continuous.
\end{proposition}
Proof: We need to prove that for every program $P$ and every 
monotonically decreasing sequence $\la X_n\ra_{n \in N}$,  
\[
\GL_P(\bigcap_{n \in N} X_n) = \bigcup_{n\in N}\GL_P(X_n).
\]
Our goal is to prove two inclusions: $\subseteq$, and $\supseteq$. \\
We first show $\supseteq$. Since 
\[
\bigcap_{j \in N}X_j \subseteq X_n
\]
for every $n \in N$, by antimonotonicity of $\GL_P$ we have
\[
\GL_P (X_n) \subseteq \GL_P(\bigcap_{j \in N} X_j). 
\]
As $n$ is arbitrary,
\[
\bigcup_{n \in N} \GL_P (X_n) \subseteq \GL_P(\bigcap_{j \in N} X_j). 
\]
Thus the inclusion $\supseteq$ holds.\\
Conversely, let $p \in \GL_P (\bigcap_{n\in N} X_n)$. Then, by
Proposition \ref{p.gl}, there must be a proof scheme $S$ with support 
support $U$ and conclusion $p$ such that  
\[
U \cap \bigcap_{n\in N} X_n = \emptyset.
\]
But the family $\la X_n \ra_{n\in n}$ is monotonically descending and
the set $U$ is finite. Thus there is an integer $n_0$ so that 
\[
U \cap X_{n_0} = \emptyset.
\]
This, however, implies that $p \in \GL_P (X_{n_0})$, and thus
\[
p \in 
\bigcup_{n \in N} \GL_P (X_n). 
\]
As $p$ is arbitrary, the inclusion $\subseteq$ holds.
Thus $\GL_P(\bigcap_{n \in N} X_n) = \bigcup_{n\in N}\GL_P(X_n)$. 
$\mbox{\  }\hfill\Box$

The  lower-half continuity of antimonotone operators is closely related to
programs, as shown in the following result.

\begin{proposition}\label{p.lhcontC}
Let $\at$ be a denumerable set of atoms. Let $f$ be an antimonotone and
lower-half continuous operator on ${\cal P}(\at )$. Then there exists 
a normal logic program $P$ such that  $f = \GL_P$. 
\end{proposition}
Proof. 

We define the program $P = P_f$ as follows:
\[
P = \{ p \lar \neg q_1,\ldots, \neg q_i : p \in f(\at \setminus \{q_1,\ldots,
q_i\})\}. 
\]
We claim that $f = \GL_P$, that is, for all $X$, $f(X) = \GL_P(X)$.

Let $X \subseteq \at$ be given. We consider two cases.\\
{\bf Case} 1: $X$ is cofinite, $X = \at \setminus \{q_1,\ldots, q_i\}$. 
We need
to prove two inclusions,  (a) $f(X) \subseteq \GL_P(X)$ and (b)
$\GL_P(X) \subseteq f(X)$.\\
For (a), note that if $p \in f(X)$, then the clause $p \lar  \neg q_1,\ldots,
\neg q_i$ belongs to $P$. Hence $p \lar$ belongs to $P_X$ and $p \in
\GL_P(X)$. 

For (b), note that if 
 $p \in \GL_P (X)$, then given the form of the clauses in $P$, 
there must be some clause $p \lar \neg q_{i_1},\ldots, \neg q_{i_j}$ in $P$ 
where 
$\{q_{i_1}, \ldots, q_{i_j} \} \subseteq \{q_1, \ldots, q_i\}$. But 
this means that $p \in f(\at \setminus \{q_{i_1}, \ldots, q_{i_j}\})$. 
Since $f$ is antimonote and $\at \setminus \{q_1, \ldots,q_i\} 
\subseteq \at \setminus \{q_{i_1}, \ldots, q_{i_j}\}$, we must have 
$$
f(\at \setminus \{q_{i_1}, \ldots, q_{i_j}\}) 
\subseteq f(\at \setminus \{q_1, \ldots,q_i\}) = f(X)$$
and, hence, $p \in f(X)$. Thus $\GL_P(X) \subseteq f(X)$. \\
\ \\
{\bf Case} 2: $X$ is not cofinite. Let $\{q_0,q_1,\ldots\}$  be an enumeration
of $\at \setminus X$. Let $Y_i = \at \setminus \{q_0,\ldots, q_i\}$. Then,
clearly, $X \subseteq Y_i$ for all $i \in N$. Moreover the sequence $\la
Y_i\ra_{i \in N}$ is monotonically decreasing  and $\bigcap_{i\in N}Y_i = X$.
Therefore, by our assumptions on the operator $f$, 
\[
f(X) = \bigcup_{i\in N} f(Y_i).
\]
Again, we need
to prove two inclusions, (a) $f(X) \subseteq \GL_P(X)$ and (b)
$\GL_P(X) \subseteq f(X)$. For (a), note that if 
$p \in f(X)$, then for some $i \in N$, $p \in F(Y_i)$. 
Therefore, for that $i$, $p \lar \neg q_0,\ldots, \neg
q_i$ is a clause in $P$. But then $X \cap \{q_0, \ldots, q_i\} = \emptyset$ 
so that  the clause $p \lar$ is in  $P_X$ and $p \in \GL_P(X)$.\\

For the proof of (b), note that if $p \in \GL_P(X)$, then because of the
syntactic form of the clauses in our program there are atoms $r_0,\ldots, r_k$
so that the clause $p \lar \neg r_0,\ldots, \neg r_k$ belongs to the program
$P$, and $r_0,\ldots, r_k \notin X$. Thus $ \{r_0,\ldots, r_k\} \subseteq
\{q_0,q_1,\ldots \}$ and, hence, for some $i \in N$, $\{r_0,\ldots, r_k\}
\subseteq \{q_0,\ldots, q_i\}$.
Now, consider such a $Y_i$. Since $Y_i$ is cofinite, it 
follows from Case 1 that $f(Y_i) = \GL_P(Y_i)$. Since $X \subseteq Y_i$, $f(Y_i) \subseteq
f(X)$ by the antimonotonicity of $f$. But $p \in \GL_P(Y_i)$ because
$r_0,\ldots, r_k \notin Y_i$ and, hence, $p \in f(Y_i)$. But 
since $f(Y_i) \subseteq f(X)$,  $p \in f(X)$ as
desired. $\hfill\Box$

We are now ready to prove the next result of this paper.

\begin{proposition}\label{p.fsp}
Let $P$ be a normal propositional program. The following are equivalent:
\begin{enumerate}
\item[$(a)$] $P$ is an {\em FSP}-program.
\item[$(b)$] The operator $\GL_P$ is upper-half continuous, i.e.
\[
\GL_P(\bigcup_{n\in N} X_n) = \bigcap_{n \in N} \GL_P (X_n)
\]
for every monotonically increasing sequence $\la X_n\ra_{n \in N}$.
\end{enumerate}
\end{proposition}
Proof: Two implications need to be proved: $(a) \Rightarrow (b)$, and
$(b) \Rightarrow (a)$.\\
Proof of the implication $(a) \Rightarrow (b)$. Here, assuming $(a)$, we
need to prove two inclusions: \\
(i) $\GL_P(\bigcup_{n\in N} X_n) \subseteq \bigcap_{n \in N} \GL_P (X_n)$,
and \\
(ii) 
$\bigcap_{n \in N} \GL_P (X_n) \subseteq \GL_P(\bigcup_{n\in N} X_n)$. \\
To prove (i), note that since $X_n \subseteq \bigcup_{j\in N} X_j$, we have
\[
\GL_P (\bigcup_{j \in N} X_j) \subseteq \GL_P (X_n).
\]
As $n$ is arbitrary,
\[
\GL_P (\bigcup_{j \in N} X_j) \subseteq \bigcap_{n\in N} \GL_P (X_n).
\]
This proves (i).\\
To prove (ii), let $p \in \bigcap_{n\in N} \GL_P (X_n)$. Then, for every
$n \in N$, $p \in \GL_P (X_n)$ and so,  for every $n \in N$, 
there is an inclusion-minimal support $U$ for $p$ such that 
\[
U \cap X_n = \emptyset .
\]
But by (a) there are only finitely many inclusion-minimal supports for
$P$-proof schemes for $p$. Therefore there is a support 
of an inclusion minimal support of a proof scheme of $p$, $U_0$,  such that for
infinitely many $n$'s
\[
U_0 \cap X_n = \emptyset .
\]
But the sequence $\la X_n\ra_{n \in N}$ is monotonically increasing.
Therefore for {\em all} $n \in N$, $U_0 \cap X_n =
\emptyset$. But then
\[
U_0 \cap \bigcup_{n \in N} X_n = \emptyset,
\]
so that $p \in GL_P( \bigcup_{n \in N} X_n)$. Thus (ii) holds and the
implication $(a) \Rightarrow (b)$ follows.\\

To prove that $(b) \Rightarrow (a)$, assume that the operator $\GL_P$ is
upper-half continuous.  We need to show that for every $p$, the reduced
defining equation for $p$ is finite. So let us assume that $\mathit{rEq}^P_p$
is not finite.  This means that there is an infinite set ${\cal X}  
= \{U_1,U_2, \ldots \}$, where $U_1 \prec U_2 \prec \cdots $, such that 
\begin{enumerate}
\item each $U_i$ is finite, 
\item the elements of $\cal X$ are pairwise inclusion-incompatible, and 
\item for every set of atoms $M$, $p \in \GL_P(M)$ if and only if
for some $U_i \in {\cal X}$, $U_i \cap M = \emptyset$.
\end{enumerate}

We will now define two sequences:
\begin{enumerate}
\item a sequence $\la K_n\ra_{n\in N}$ of infinite sets of integers and 
\item a sequence $\la p_n \ra_{n \in N \setminus \{0\}}$ of atoms.
\end{enumerate}
We define $K_0 = N$, and we define $p_1$ as the first element of $U_1$
such that
\[
\{ j : p \notin U_j\}
\]
is infinite. Clearly, $K_0$ is well-defined. We need to show that $p_1$
is well-defined. If $p_1$ is not well-defined, then for every $p \in U_1$
there is an integer $i_p$ such that for all $m > i_p$, $p \in U_m$. But
$U_1$ is finite so taking $n = \max_{p \in U_1} i_p$, we find that
for {\em all} $m >  n$, $U_1 \subseteq U_m$ - which contradicts the fact
that the sets in $\cal X$ are pairwise inclusion-incompatible.
Thus $p_1$ is well-defined. We now set 
\[
K_1 = \{ n \in K_0 : p_1 \notin U_n\} =\{n \in K_0 : \{p_1\} \cap U_n = \emptyset\}.
\]
Clearly. $K_1$ is infinite.

Now, let us assume that we already defined $p_l$ and $K_l$ so that 
$K_l = \{n : U_n \cap \{p_1,\ldots, p_l\} = \emptyset\}$ is an infinite 
subset of $N$. 
We select $p_{l+1} $  as the first element $p \in U_{l+1}$ so that
\[
\{ j : j \in K_l \ \mbox{and} \  p \notin U_j\}
\]
is infinite. Clearly, by an argument as above, there is such $p$, and so
$p_{l+1}$ is well-defined.  We then set
\[
K_{l+1} = \{j \in K_l : p_{l+1} \notin U_j\}.
\]
Since $\{p_1,\ldots, p_l\} \cap U_j = \emptyset$ for all $j \in K_l$,
$\{p_1,\ldots, p_{l+1}\} \cap U_j = \emptyset$ for all $j \in K_{l+1}$.
By construction, the set $K_{l+1}$ is infinite.

Now, we complete the argument as follows. We set $X_n = \{p_1, \ldots,
p_n\}$. The sequence $\la X_n\ra_{n \in N}$ is monotonically increasing.
For each $n$ there is $j$ (in fact infinitely many $j$'s) so that
$X_n \cap U_j = \emptyset$. Therefore, for each $n$, $p \in \GL_P
(X_n)$. Hence $p \in \bigcap_{n \in N} \GL_P(X_n)$.

On the other hand, let $X = \bigcup_{n \in N} X_n$. Then
\[
X = \{p_1,p_2,...\}.
\]
By our construction, $p_n \in U_n$, and so $U_n \cap X \neq \emptyset$.
Therefore $X$ does not admit {\em any} $P$-proof scheme for $p$. Thus $p
\notin \GL_P(X) = \GL_P(\bigcup_{n\in N} X_n)$. But this would 
contradict our assumption that $\GL_P$ is upper-half continuous. 
Thus there can be no such $p$ and hence $P$ must be a {\em FSP}-program. 
$\hfill\Box$

\section{Extensions to $\cc$-programs}\label{extCC}

In \cite{nss02} Niemel\"a and coauthors defined a significant extension
of logic programming with stable semantics which allows for 
programming with cardinality constraints, and, more generally, with
weight constraints.  This extension has been further studied in
\cite{mr04,mnt07}. To keep things simple, we will limit our discussion to
cardinality constraints only, although it is possible to extend our 
arguments to any class of convex constraints \cite{lt05}. 
{\em Cardinality constraints} are expressions of the form $lXu$, 
where $l, u \in N$, $l
\le u$ and $X$ is a finite set of atoms. The semantics of an 
atom $lXu$ is that a set of atoms $M$ satisfies $kXl$ if and only if $k \leq |M
\cap X|$.  When $l = 0$,  we do not write it, and, likewise, when $u \ge |X|$,
we omit it, too. Thus an atom $p$ has the same meaning as $1\{p\}$  while $\neg
p$ has the same meaning as $\{p\}0$.

The stable semantics for $\cc$-programs 
is defined via fixpoints of an analogue of the Gelfond-Lifschitz 
operator $GL_P$; see the details in \cite{nss02} and \cite{mr04}. The operator
in question is neither monotone nor antimonotone. But when we limit our
attention to the programs $P$ where clauses have the property that
the head consists of a single atom (i.e. are of the form $1\{p\}$), then
one can define an operator $\CCGL_P$ which is antimonotone and whose fixpoints
are stable models of $P$. This is done as follows.\\
\ \\
Given a clause $C$
\[
p \leftarrow l_1X_1u_1,\ldots , l_mX_mu_m,
\]
we transform it into the clause
\begin{equation}\label{transf}
p \leftarrow l_1X_1,\ldots , l_mX_m, X_1u_1,\ldots ,X_mu_m
\end{equation}
\cite{mnt07}.
We say that a clause $C$ of the form (\ref{transf}) is a $\cc$-Horn 
clause if it is of the form 
\begin{equation}\label{CChorn}
p \leftarrow l_1X_1,\ldots , l_mX_m.
\end{equation}
A $\cc$-Horn program is a $\cc$-program all of whose clauses are of 
the form (\ref{CChorn}). If $P$ is a $\cc$-Horn program, we 
can define the analogue of the one step provability operator $T_P$ by 
defining that for a set of atom $M$, 
\begin{equation}\label{CConestep}
T_P(M) = \{p: (\exists C = p \leftarrow l_1 X_1,\ldots , l_mX_m) 
(\forall i \in \{1, \ldots m\}) ( |X_i \cap M| \geq l_i)\}
\end{equation}
It is easy to see that $T_P$ is monotone operator that 
the least fixed point of $T_P$ is given by 
\begin{equation}\label{CClfp}
\lfp(T_P) = \bigcup_{n \geq 0} T_P^n(\emptyset).
\end{equation}

We can define the analogue of the Gelfond-Lifschitz reduct of a $\cc$-program, 
which we call the $\mathit{NSS}$-reduct of $P$, as follows. 
Let $\bar{P}$ denote the set of all transformed clauses derived from 
$P$. Given a set of atoms $M$, we eliminate from $\bar{P}$ those clauses 
where some  upper-constraint ($X_iu_i$) is  not satisfied by $M$, 
i.e. $|M \cap X_i| > u_i$.   
In the remaining clauses, the constraints of the form $X_iu_i$ are eliminated
altogether. This leaves us with a $\cc$-Horn program $P_M$. 
We then define $\CCGL_P(M)$ to be the least fixed point 
of $T_{P_M}$ and say that $M$ is a $\cc$-stable model if 
$M = \CCGL_P(M)$. 
The equivalence of this construction and the original construction 
in \cite{nss02} for normal $\cc$-programs is shown in \cite{mnt07}. 

Next we define the analogues of $P$-proof schemes for normal 
$\cc$-programs, i.e. programs which consists entirely of clauses 
of the form (\ref{transf}). This is done by induction as follows. When
\[
C = p \leftarrow X_1u_1,\ldots, X_ku_k
\]
is a normal $\cc$-clause without the cardinality-constraints of the form
$l_iX_i$ then
\[
\la \la C, p \ra, \{X_1u_1,\ldots, X_ku_k\} \ra
\]
is a $P$-$\cc$-proof scheme with support $\{X_1u_1,\ldots, X_ku_k\}$.
Likewise, when
\[
S = \la \la C_1,p_1\ra,\ldots, \la C_n , p_n \ra, U \ra
\]
is a $P$-$\cc$-proof scheme, 
\[
p \leftarrow l_1X_1,\ldots , l_mX_m, X_1u_1,\ldots ,X_mu_m
\]
is a clause in $P$, and $|X_1 \cap \{p_1,\ldots, p_n\}| \ge l_1$,
$\ldots$, 
$|X_m \cap \{p_1,\ldots, p_n\}| \ge l_m$, then
\[
\la \la C_1,p_1\ra,\ldots, \la C_n , p_n \ra, \la C, p\ra , U \cup
\{X_1u_1,\ldots, X_mu_m\} \ra
\]
is a $P$-$\cc$-proof scheme with support $U \cup
\{X_1u_1,\ldots X_mu_m\}$.
The notion of admittance of a $P$-$\cc$-proof scheme is similar to the notion 
of admittance of $P$-proof scheme for normal programs $P$. That is, 
if $ \mathcal{S} = \la \la C_1,p_1\ra,\ldots, \la C_n , p_n \ra, \la C, p\ra ,
U\ra$ is a $\cc$-proof scheme with support $U = \{X_1u_1, \ldots X_n u_n \}$,
then $\mathcal{S}$ is admitted by $M$ if for every  
$X_iu_i \in U$, $M \models X_iu_i$, i.e. $|M\cap X_i| \le u_i$.

Similarly, we can associate a propositional formula 
$\phi_U$ so that $M$ admits $\mathcal{S}$ if and only if 
$M \models \phi_U$ as follows:
\begin{equation}\label{eq:CCsupp}
\phi_U = \bigwedge_{i=1}^n \bigvee_{W \subseteq X_i,|W| = |X_i| -u_i} 
\neg W.
\end{equation}
Then we can define a partial ordering on the set of possible 
supports of proof scheme by defining 
$U_1 \preceq U_2 \iff \phi_{U_2} \models \phi_{U_1}$.  
For example if $U_1 = \la \{1,2,3\} 2$, $\{4,5,6\}2 \ra$ and 
$U_2 = \la \{1,2,3,4,5,6\},4 \ra$, then \\
\begin{eqnarray*}
\phi_{U_1} &=& (\neg 1 \vee \neg 2 \vee \neg 3) \wedge (\neg 4 \vee \neg 5 
\lor \neg 6) \\
\phi_{U_2} &=& \bigvee_{1 \leq i < j \leq 6} (\neg i \wedge \neg j).
\end{eqnarray*}
Then clearly $\phi_{U_1} \models \phi_{U_2}$ so that 
$U_2 \preceq U_1$. 
We then define a normal propositional $\cc$-program to be 
{\em FPS $\cc$-program} if for each $p \in At$, there 
are finitely many $\preceq$-minimal supports of $P$-$\cc$-proof schemes 
with conclusion $p$.

We can also define analogue of the defining equation $CCEq^P_p$ of 
$p$ relative to a normal $\cc$-program $P$ as 
\begin{equation}\label{CCdefine}
p \Leftrightarrow (\phi_{U_1} \lor \phi_{U_2} \lor \cdots )
\end{equation}
where $\la U_1, U_2, \ldots \ra$ is a list of supports of 
all $P$-$\cc$-proofs schemes with conclusion $p$. 
Again up to a total ordering of possible finite supports, 
this formula is unique. Let $\Phi_P$ be the set 
$\{CCEq^P_p: p \in At\}$. Similarly, we define the {\em reduced
defining equation} for $p$ relative to $P$ to be the formula
\begin{equation}\label{reddefeqCC}
p \Leftrightarrow (\neg \phi_{U_1} \lor \neg \phi_{U_2} \lor \ldots )
\end{equation}
where $U_i$ range over {\em $\preceq$-minimal} supports of $P$-$\cc$-proof
schemes for the atom $p$. 

Then we have the following analogues 
of Propositions \ref{p.char}  and \ref{p.eq}. 

\begin{proposition}\label{CCp.char}
For every normal propositional $\cc$-program $P$ and every set $M$ of
atoms, $M$ is a $\cc$-stable model of $P$ if and only if the 
following two conditions hold:
\begin{enumerate}
\item[(i)] for every $p \in M$,  there is a $P$-$\cc$-proof scheme $S$ with
conclusion $p$ such
that $M$ admits $S$ and 
\item[(ii)] for every $p \notin M$,  there is no $P$-$\cc$-proof scheme $S$ 
with conclusion $p$ such that $M$ admits $S$.
\end{enumerate}
\end{proposition}

\begin{proposition}\label{CCp.eq}
Let $P$ be a normal propositional $\cc$-program.
Then $\cc$-stable models of $P$ are precisely the propositional models of the
theory $\Phi_P$.
\end{proposition}

We also can prove the analogues of Propositions \ref{p.anti}
and \ref{p.gl}.

\begin{proposition}\label{CCp.gl88} 
For any $CC$-program $P$, the operator $\CCGL_P$ is antimonotone.
\end{proposition}
Proof: 
It is easy to see that if $M_1 \subseteq M_2$, then for any clause 
\[
C= p \rightarrow l_1 X_1, \ldots, l_m X_m, X_1 u_1, \ldots X_m l_m,
\]
$M_2 \models X_i u_i$ implies $M_1 \models X_i u_i$. Thus 
it follows that $P_{M_2} \subseteq P_{M_1}$ and hence 
$\lfp(T_{P_{M_2}}) \subseteq \lfp(T_{P_{M_1}})$. 
$\hfill\Box$ 

\begin{proposition}\label{CCp.gl}
Let $P$ be a normal propositional $\cc$-program and $M$ be a set of
atoms. Then 
\begin{multline*}
\CCGL_P(M) = \{p : \ \mbox{there exists a $P$-proof scheme $S$}
\ \mbox{ such that $M$ admits $S$},  \\
\mbox{and $p$ is the conclusion of $S$}\}
\end{multline*}
\end{proposition}
Proof: Let us assume that  $p \in \CCGL_P(M)$, i.e. $p \in \lfp(T_{P_M})$. 
Since $\lfp(T_{P_M}) = \bigcup_{n \geq 1} T_{P_M}^n (\emptyset )$, we 
can easily show by induction on $n$ that if 
$p \in T_{P_M}^n (\emptyset )$, then there is a
$P$-$\cc$-proof
scheme $S_p$ such  $p$ is the conclusion of $S_p$  and $S_p$ is
admitted by $M$.\\
Conversely, we can show,  
by induction on the length of the $P$-$\cc$-proof schemes, 
that whenever there is $P$-$\cc$-proof scheme $S$ admitted by $M$, then $p$
belongs to $\lfp(T_{P_M})$. $\mbox{  }\hfill\Box$

Next we prove that analogue of Proposition \ref{p.lhcont}.

\begin{proposition}\label{CCp.lhcont}
For every normal $\cc$-program $P$, the operator $\CCGL_P$ is lower-half
continuous.
\end{proposition}
Proof: We need to prove that for every normal $\cc$-program $P$ and every 
monotonically decreasing sequence $\la X_n\ra_{n \in N}$ 
\[
\CCGL_P(\bigcap_{n \in N} X_n) = \bigcup_{n\in N}\CCGL_P(X_n).
\]
We need to prove two inclusions: $\subseteq$, and $\supseteq$. \\
We first show $\supseteq$. Since 
\[
\bigcap_{j \in N}X_j \subseteq X_n
\]
for every $n \in N$, it follows from the antimonotonicity of $\CCGL_P$ that 
we have
\[
\CCGL_P (X_n) \subseteq \GL_P(\bigcap_{j \in N} X_j). 
\]
As $n$ is arbitrary,
\[
\bigcup_{n \in N} \CCGL_P (X_n) \subseteq \CCGL_P(\bigcap_{j \in N} X_j). 
\]
Thus the inclusion $\supseteq$ holds.\\
Conversely, let $p \in \CCGL_P (\bigcap_{n\in N} X_n)$. Then, by
Proposition \ref{CCp.gl}, there must be a $\cc$-proof scheme $S$ with support 
support $U=\{Y_1 u_1, \ldots, Y_n u_n\}$ and conclusion $p$ such that  
\[
|Y_i \cap \bigcap_{n\in N} X_n| \leq u_i \mbox{ for } i =1, \ldots, n.
\]
Since the family $\la X_n \ra_{n\in n}$ is monotonically descending, it 
follows that 
$$ Y_i \cap X_1 \supseteq Y_i \cap X_2 \supseteq \cdots.$$
Since $Y_i$ is finite, it is the case that if 
$|Y_i \cap \bigcap_{n\in N} X_n| \leq u_i$, 
then there is some $m_i$ such that 
$|Y_i \cap X_{m_i}| \leq u_i$. Hence if $m = \max (m_1, \ldots, m_n)$, then 
\[
|Y_i \cap X_m| \leq u_i \mbox{ for } i =1, \ldots, n.
\]
 This, however, implies that $p \in \CCGL_P (X_{m})$, and thus
\[
p \in 
\bigcup_{n \in N} \CCGL_P (X_n). 
\]
As $p$ is arbitrary, the inclusion $\subseteq$ holds.
Thus $\CCGL_P(\bigcap_{n \in N} X_n) = \bigcup_{n\in N}\CCGL_P(X_n)$. 
$\hfill\Box$

Next we can prove the analogue of the first half of Proposition 
\ref{p.fsp}. 

\begin{proposition}\label{CCp.fsp}
Let $P$ be a normal propositional $\cc$-program. Then if 
$P$ is an {\em FSP}-program, the operator $\CCGL_P$ is upper-half continuous,
i.e.  
\[
\CCGL_P(\bigcup_{n\in N} X_n) = \bigcap_{n \in N} \CCGL_P (X_n)
\]
for every monotonically increasing sequence $\la X_n\ra_{n \in N}$.
\end{proposition}
Proof: Two implications need to be proved: $(a) \Rightarrow (b)$, and
$(b) \Rightarrow (a)$.\\
Proof of the implication $(a) \Rightarrow (b)$. Here, assuming $(a)$ we
need to prove two inclusions: \\
(i) $\GL_P(\bigcup_{n\in N} X_n) \subseteq \bigcap_{n \in N} \GL_P (X_n)$,
and \\
(ii) 
$\bigcap_{n \in N} \GL_P (X_n) \subseteq \GL_P(\bigcup_{n\in N} X_n)$. \\
To prove (i), note that since $X_n \subseteq \bigcup_{j\in N} X_j$, we have
\[
\CCGL_P (\bigcup_{j \in N} X_j) \subseteq \CCGL_P (X_n).
\]
As $n$ is arbitrary,
\[
\CCGL_P (\bigcup_{j \in N} X_j) \subseteq \bigcap_{n\in N} \CCGL_P (X_n).
\]
This proves (i).\\
To prove (ii), let $p \in \bigcap_{n\in N} \CCGL_P (X_n)$. Then, for every
$n \in N$, $p \in \CCGL_P (X_n)$ and so,  for every $n \in N$, 
there is a minimal  support $U_n=\{Y_1^{(n)} u_1^{(n)}, \ldots, Y_{m_n}^{(n)}
u_{M_n}^{(n)}\}$ for $p$ such that \[
|Y_i^{(n)} \cap X_n| \leq u_i^{(n)} \mbox{ for } i = 1, \ldots, m_n.
\]

But  there are only finitely many $\preceq$-minimal supports for
$P$-$\cc$-proof schemes for $p$. Therefore there is a support 
$U_0= \{Z_1 w_1, \ldots, Z_t w_t\}$ for a $P$-$\cc$-proof scheme with 
conclusion $p $ such that for
infinitely many $n$'s
\[
|Z_i \cap X_n| \leq w_i \mbox{ for } i =1, \ldots, t.
\]
But the sequence $\la X_n\ra_{n \in N}$ is monotonically increasing.
Therefore for {\em all} $n \in N$, 
\[
|Z_i \cap X_n| \leq w_i \mbox{ for } i =1, \ldots, t.
\]
But since each $Z_i$ is finite, then it must be the case that 
\[
|Z_i \cap \bigcup_{n in N} X_n| \leq w_i \mbox{ for } i =1, \ldots, t.
\]
so that $p \in \CCGL_P( \bigcup_{n \in N} X_n)$. 
$\hfill\Box$

We note that, alternatively,  one can easily give a direct reduction of our
$\cc$-programs to normal logic programs using the methods of
\cite{fl05} and  the distributivity result of \cite{ltt99}. Such reduction, 
of course, lead to an exponential blow up in the size of the representation.

\section{Conclusions}\label{concl}
We note that investigations of proof systems in a related area, SAT,  
play a key role in establishing  lower bounds on the complexity of algorithms
for finding the models. We wonder if there are analogous results in ASP. For
achieving such a goal,  we need to find and investigate proof systems for
ASP. One candidate for such a proof system is provided in this paper 
by using $P$-proof schemes. 
We wonder if such a proof system can be used to develop a deeper understanding
of the complexity issues related to finding stable models.

\section*{Acknowledgments}

This research of the first author was supported by the National Science
Foundation under Grant IIS-0325063. 
This research of the second author was supported by the National Science
Foundation under Grant DMS 0654060.

\bibliographystyle{plain}

\end{document}